\documentclass[9pt,journal,compsoc]{IEEEtran}
\usepackage[utf8]{inputenc}
\usepackage{algorithm2e}
\usepackage{tikz}
\usepackage{pgfplots}
\usepackage{subcaption}
\usepackage{graphicx}
\usepackage{csvsimple}
\usepackage[colorinlistoftodos]{todonotes}
\title{Generalized Convolutional Neural Networks for Point Cloud Data}
\date{January 2017}
\author{ \parbox{2 in}{\centering Aleksandr Savchenkov \\
        {\tt\small asavchenkov1@gmail.com \newline Cylance Inc.}}
        \parbox{2 in}{ \centering Andrew Davis \\
        {\tt\small adavis@cylance.com \newline Cylance Inc. Irvine, CA 92617}}
        \parbox{2 in}{ \centering Xuan Zhao \\
        {\tt\small xzhao@cylance.com \newline Cylance Inc.}}
}
\usepackage[numbers]{natbib}
\usepackage{graphicx}

\begin{document}

\maketitle

\begin{abstract}
    The introduction of cheap RGB-D cameras, stereo cameras, and LIDAR devices has given the computer vision community 3D information that conventional RGB cameras cannot provide. This data is often stored as a point cloud, i.e. a set of 3D points. In this paper, we present a novel method to apply the concept of convolutional neural networks to this type of data. By creating a mapping of nearest neighbors in a dataset, and individually applying weights to spatial relationships between points, we achieve an architecture that works directly with point clouds, but closely resembles a convolutional neural net in both design and behavior. Such a method bypasses the need for extensive feature engineering, while proving to be computationally efficient and requiring few parameters.
\end{abstract}

\section{Introduction and Related Work}
Over the past half decade, sensors capable of precisely measuring distances have dropped in price dramatically. RGB-D (RGB + Distance) cameras such as the Microsoft Kinect are able to assign distances to individual pixels, and LIDAR (Light Detection and Ranging) scanners are more effective and affordable. A combination of these advances in hardware and research into SLAM (Simultaneous Localization and Mapping) have allowed robots and self driving cars to stitch together individual images into maps of their environment.

Whereas 2D image based object detection and segmentation has seen plenty of advancement, the processing of point cloud data is still slightly lagging. This can be attributed partly to the ubiquity of 2D images and relative scarcity of point cloud data, but also partly to the convenient nature of RGB images, as spatial relationships between pixels are encoded in the structure of the image itself by the indices of pixels in the matrix. Convolutional Neural Networks (CNNs) exploit this efficiently, as individual pixels can be matched with sets of weights, resulting in a computationally cheap operation. In a point cloud however, individual points can exist in any location in the array, and spatial information is encoded explicitly alongside other information. A map generated from an RGB-D camera would consist of points that would each be structured as such: [X,Y,Z,R,G,B].

Current leading solutions generally involve two different classes, both of which involve converting the point cloud into a format more convenient for conventional CNNs. One way to do so involves converting the point cloud into a three dimensional grid where each value in the grid represents the presence or number of points in that volume, and applying a 3-dimensional convolution \citep{DBLP:journals/corr/HegdeZ16}. This works reasonably well, but has the downside of densifying sparse data. In addition, the size of the input grows by the cube of the granularity required of the network. The other dominant design is to project the model into multiple 2D views of the object \citep{su15mvcnn}. Ensembles of the former are currently the most effective\citep{DBLP:journals/corr/BrockLRW16}, at least with regard to the ModelNet40 and ModelNet10 datasets\citep{DBLP:journals/corr/WuSKTX14}. These are benchmark datasets of 3D models broken up into 40 and 10 classes, and with 12311 and 4917 samples respectively.

Recent work has demonstrated strong results without requiring this conversion, all the while dealing with the data natively. PointNets\citep{PointNets} calculate features for all individual points, while Kd-Networks\citep{DBLP:journals/corr/KlokovL17} split point clouds into Kd-trees, with each layer working on the features of the nodes below it.

Our new method is related to both classes. It is a convolution in a traditional mathematical sense, however it is also defined over a continuous input space, allowing it to work natively with both point clouds and images. In this manner, our technique generalizes CNNs. This bridges the gap between the former traditional CNN based models, and the latter techniques that forgo convolution in order to natively consume point clouds.

\begin{figure*}[ht!]
    \centering
    \begin{subfigure}[t]{0.49\textwidth}
        \centering
        \includegraphics[width=\linewidth]{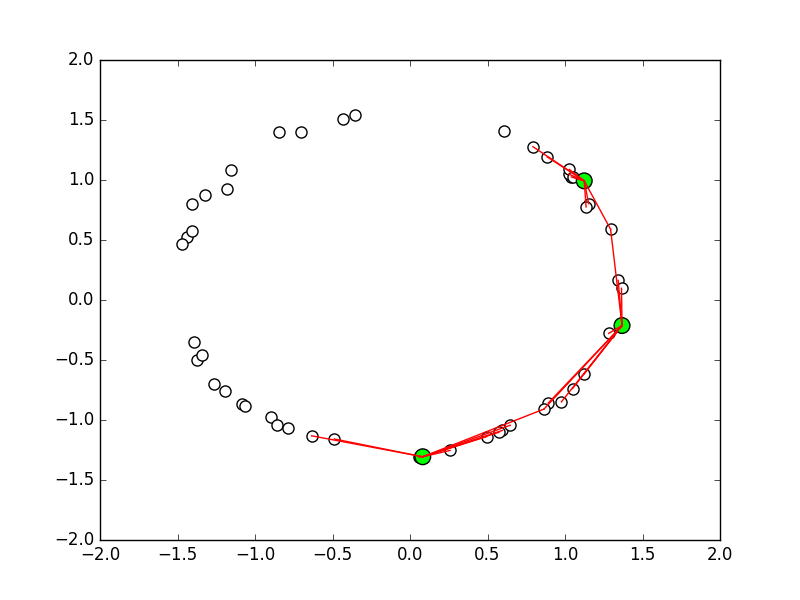}
    \end{subfigure}
    \begin{subfigure}[t]{0.49\textwidth}
        \centering
        \includegraphics[width=\textwidth]{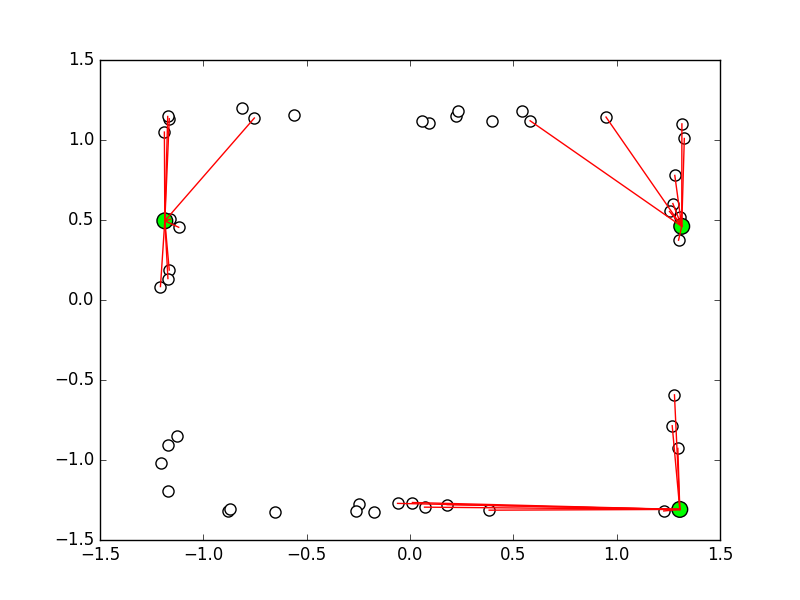}
    \end{subfigure}
    \caption{samples from the toy dataset of squares and circles with the 10 nearest neighbors shown for three points in each figure.}
\end{figure*}

\begin{figure*}[ht!]
    \centering
    \begin{subfigure}[t]{0.21\textwidth}
        \centering
        \includegraphics[width=\linewidth]{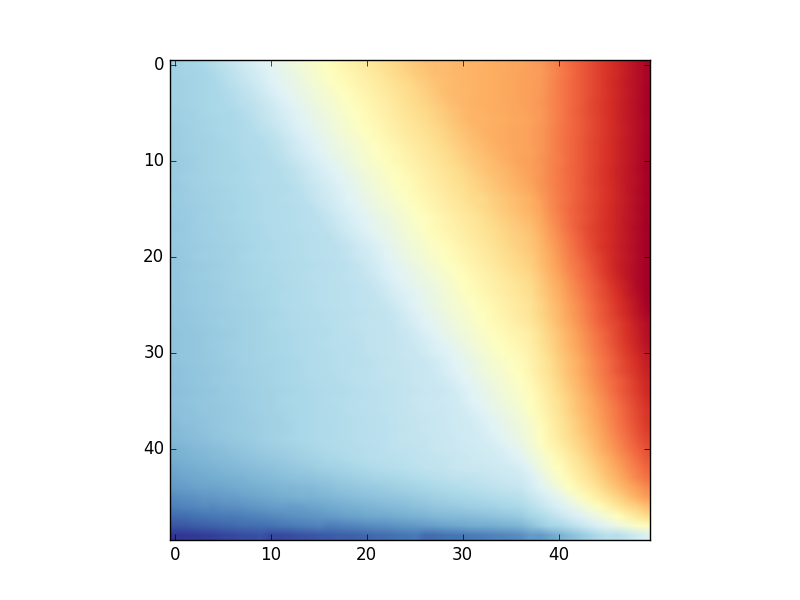}
    \end{subfigure}
    \begin{subfigure}[t]{0.21\textwidth}
        \centering
        \includegraphics[width=\linewidth]{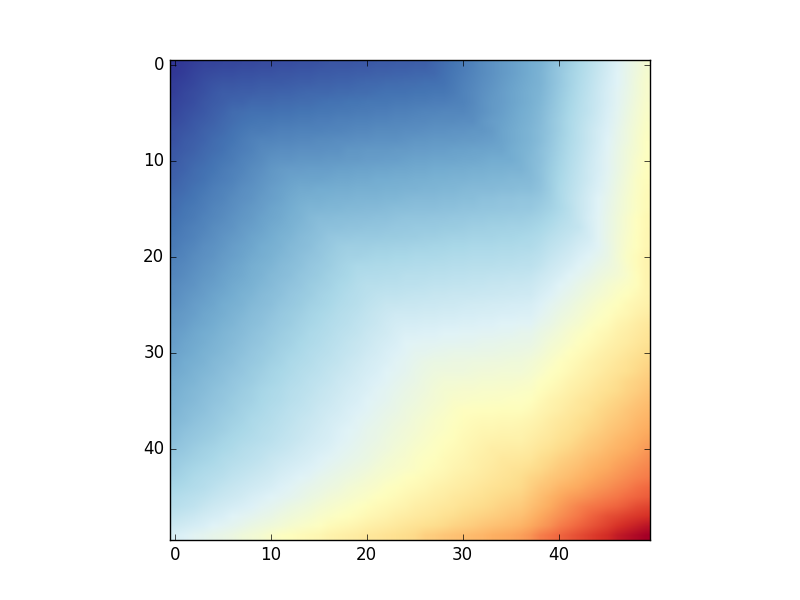}
    \end{subfigure}
    \begin{subfigure}[t]{0.21\textwidth}
        \centering
        \includegraphics[width=\linewidth]{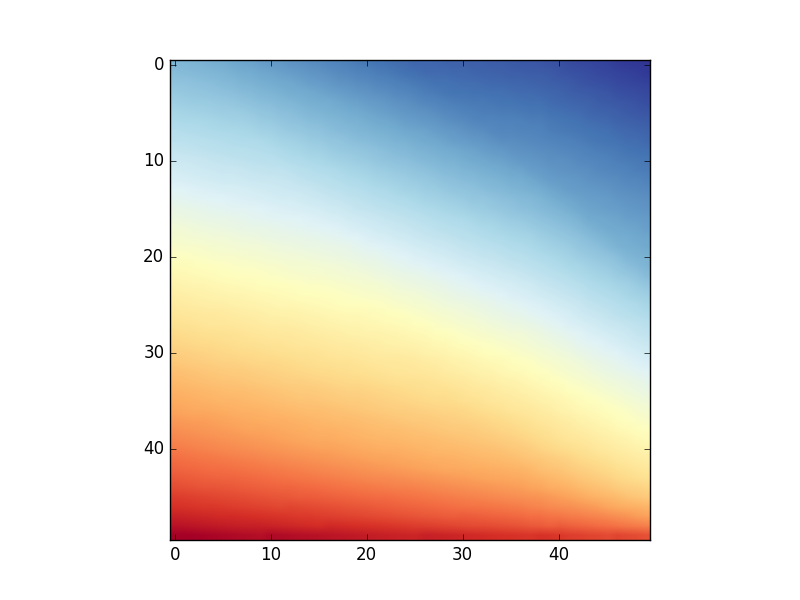}
    \end{subfigure}
    \begin{subfigure}[t]{0.21\textwidth}
        \centering
        \includegraphics[width=\linewidth]{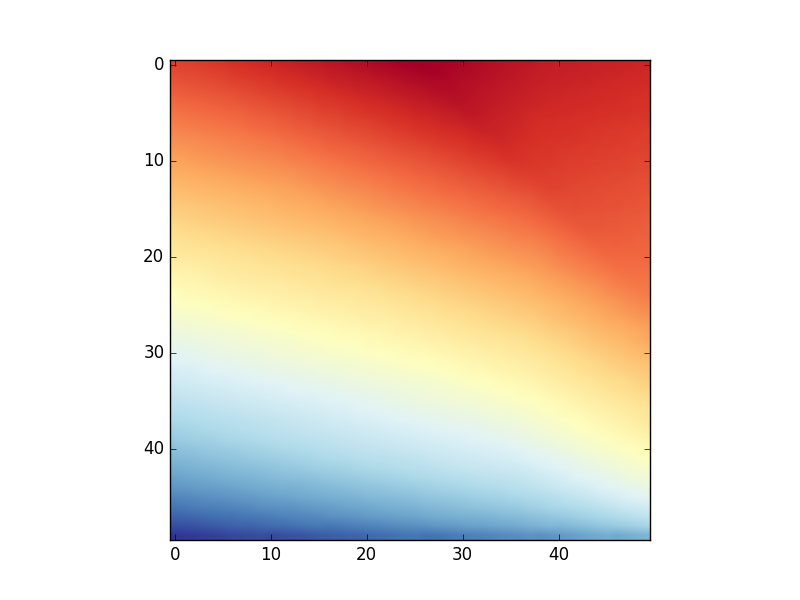}
    \end{subfigure}
    \begin{subfigure}[t]{0.05\textwidth}
        \centering
        \includegraphics[width=\linewidth]{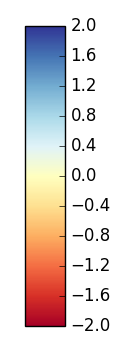}
    \end{subfigure}
    \caption{The 4 filters learned by the toy network.}
\end{figure*}

\begin{figure*}[ht!]
    \centering
    \begin{subfigure}[t]{0.2\textwidth}
        \centering
        \includegraphics[width=\linewidth]{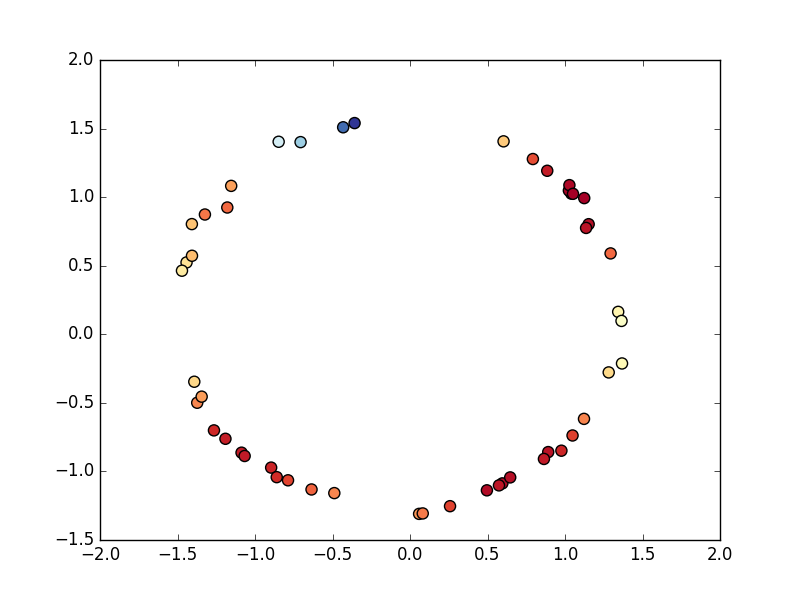}
    \end{subfigure}
    \begin{subfigure}[t]{0.2\textwidth}
        \centering
        \includegraphics[width=\linewidth]{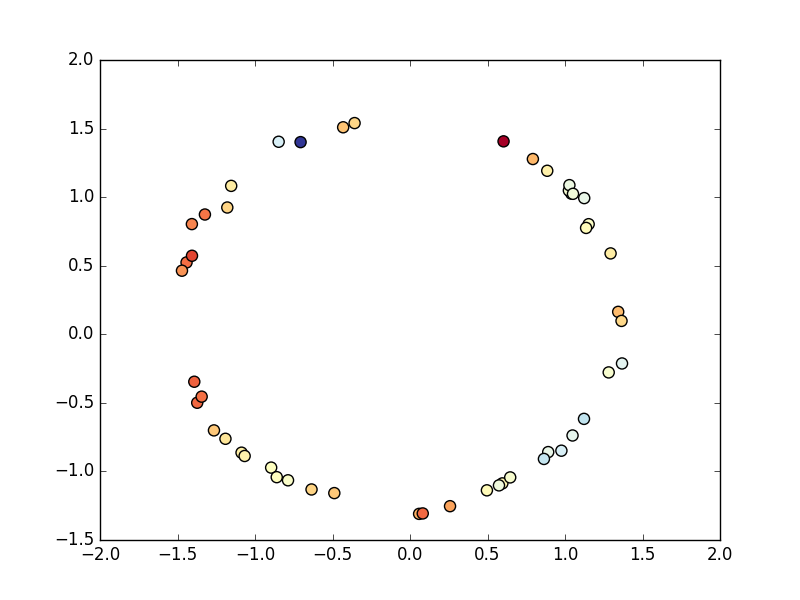}
    \end{subfigure}
    \begin{subfigure}[t]{0.2\textwidth}
        \centering
        \includegraphics[width=\linewidth]{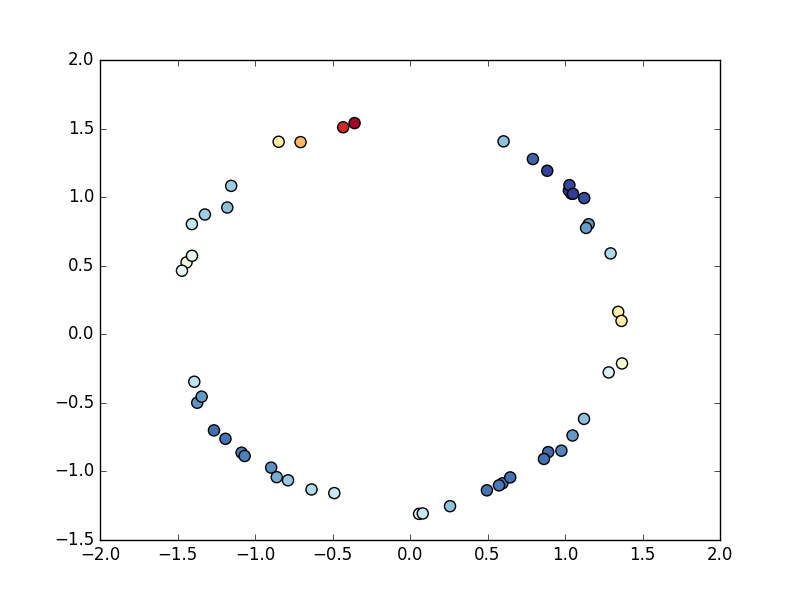}
    \end{subfigure}
    \begin{subfigure}[t]{0.2\textwidth}
        \centering
        \includegraphics[width=\linewidth]{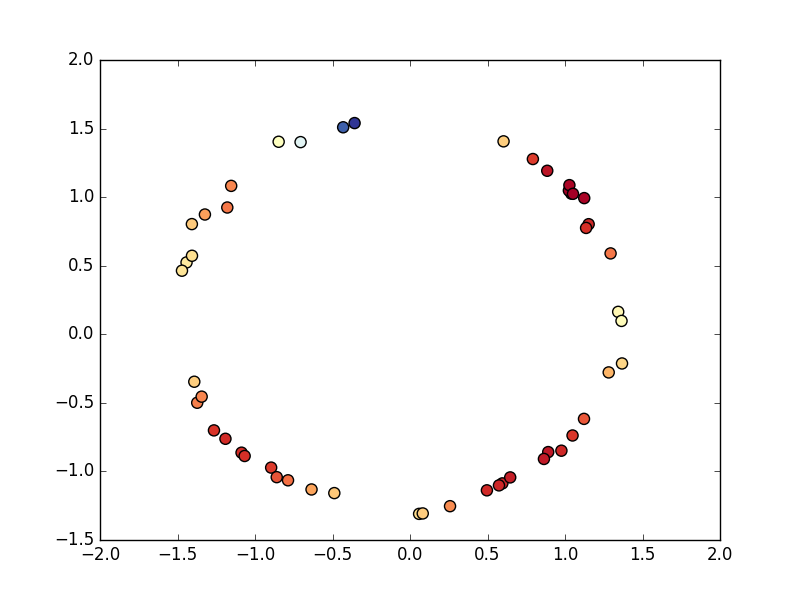}
    \end{subfigure}
    \begin{subfigure}[t]{0.05\textwidth}
        \centering
        \includegraphics[width=\linewidth]{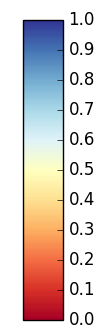}
    \end{subfigure}
    \begin{subfigure}[t]{0.2\textwidth}
        \centering
        \includegraphics[width=\linewidth]{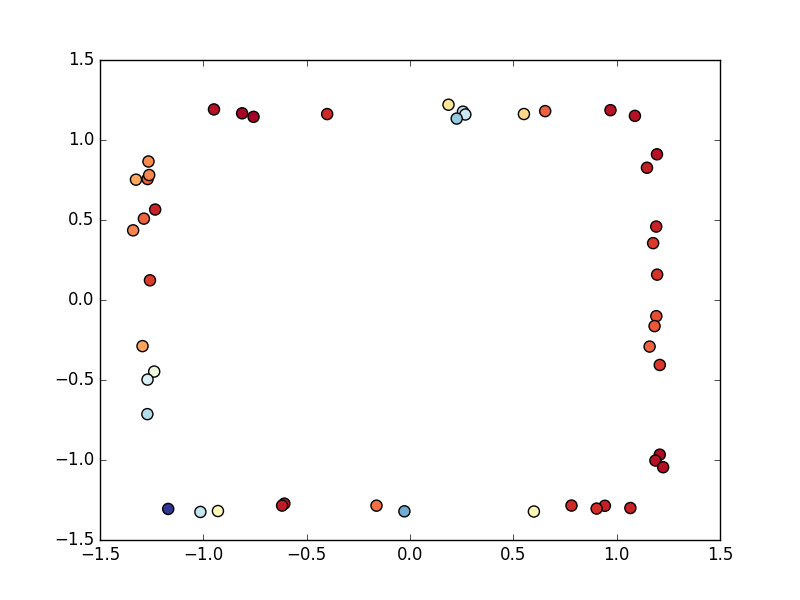}
    \end{subfigure}
    \begin{subfigure}[t]{0.2\textwidth}
        \centering
        \includegraphics[width=\linewidth]{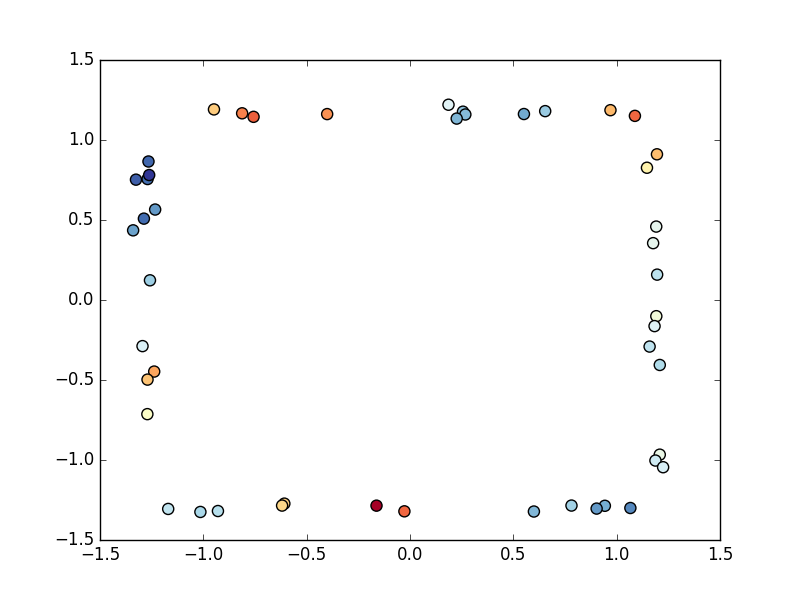}
    \end{subfigure}
    \begin{subfigure}[t]{0.2\textwidth}
        \centering
        \includegraphics[width=\linewidth]{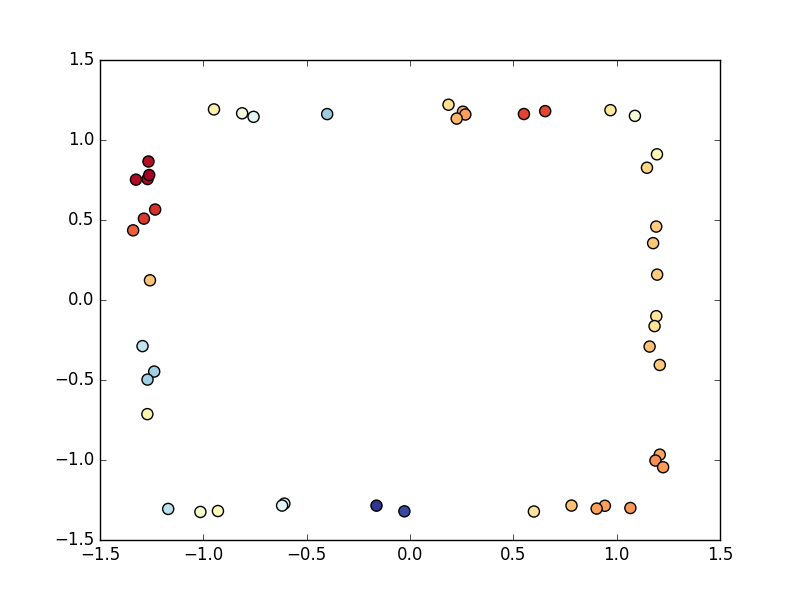}
    \end{subfigure}
    \begin{subfigure}[t]{0.2\textwidth}
        \centering
        \includegraphics[width=\linewidth]{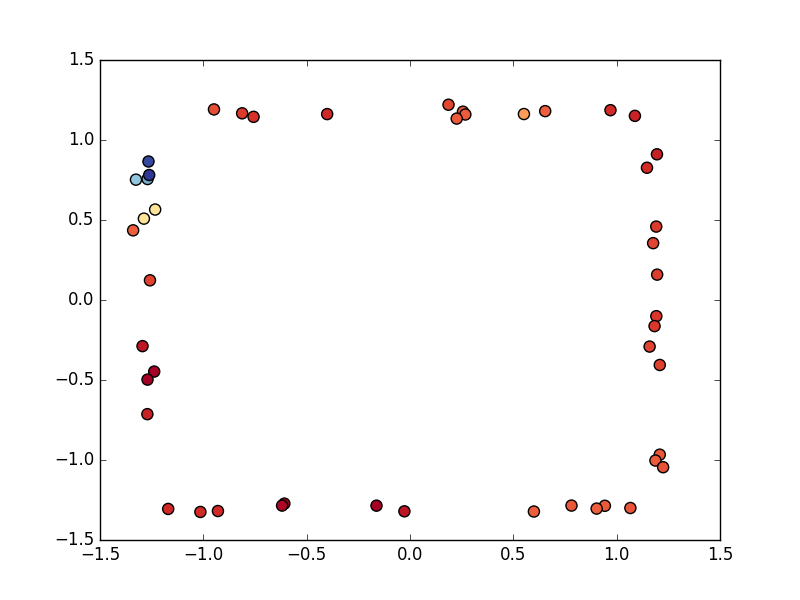}
    \end{subfigure}
    \begin{subfigure}[t]{0.05\textwidth}
        \centering
        \includegraphics[width=\linewidth]{activation_color_bar}
    \end{subfigure}
    \caption{Activations corresponding to the filters above.}
\end{figure*}

\section{Architecture}

The architecture presented not only consumes point clouds natively, but also produces point clouds in the same format, making it useful in the same exact ways and circumstances as conventional CNNs. This new layer also closely mirrors conventional CNN layers in that it involves a shared filter that is applied only locally. These filters can also be visualized, albeit with a more generalized technique that we describe later, and will bear resemblance to filters learned in 2D CNNs, with the first layer commonly performing tasks such as edge detection.

\begin{algorithm}
\SetAlgoLined
\LinesNumbered
\KwData{Matrix $M$ with shape $(S+D)*N$ \;
$S$: The number of spatial features (XYZ), 2 for our toy dataset, 3 for ModelNet10. \;
$D$: The number of conventional features (RGB), 1 for our datasets. \;
$N$: The number of points in the point cloud.}
\KwData{Small neural network $f$ with output shape $D'$}
\KwResult{Matrix $A$ with shape $(S+D')*N$}
 Extract $K$ nearest neighbors from $M$ into tensor $C$ of shape $(S+D)*N*K$\;
 $C=C-M$ for range [0:$S$,:,:] (we leave the regular features alone) \;
 $Z_{i} = \sum_{j=0}^{K} f(C_{ij})$\;
 $A = \sigma(Z)$ where $\sigma$ is an activation function of our choice\;
 
\caption{Simple Implementation of a Generalized Convolutional Layer}
\end{algorithm}

For the sake of illustration, we will demonstrate a network with one generalized convolution, one fully connected layer, and a simplified two dimensional problem as an example. It is given the task of classifying 2d point clouds as squares or circles. While this problem is trivial, it gives us an understanding of the behavior of this architecture. We present results on the ModelNet10 dataset later on.

In Algorithm 1, Line 1, we compute the indices of the k nearest neighbors to each point. Next on line 2 we then calculate spatial relationships between each query point and its neighbors. Currently we extract differences in the x,y, and in 3D scenarios z values. We also add Euclidean distance as it is already calculated during the nearest neighbor calculation. The network learns more complex features on its own. It should be noted that other features can be extracted if they prove to be useful. These relative spatial features replace absolute spatial features. After this operation we have a tensor with the shape [$N_{points}$, $N_{neighbors}$, $N_{spatial features}$+ $N_{conventional features}$].

We then apply a small neural network to each point->neighbor relationship. This works effectively as a convolution, except the effect of spatial relationships is determined through function approximation by the neural network instead of being explicitly encoded in a matrix. These output features are then summed along the neighbor dimension and run through an activation function of our choice, giving us a tensor of shape [$N_{points}$, $N_{features}$]. We then concatenate the absolute XYZ values of each point to its corresponding activations for the next layer to use for spatial feature extraction, giving us a final tensor shaped [$N_{points}$, $N_{new conventional features}$ + $N_{spatial features}$]. where $s$ is the spatial dimensionality of our original point cloud (generally either 2 or 3). In the simplest implementation, this is a new point cloud with points in all of the original locations, but storing new data. This point cloud can once again be run through the same kind of layer to extract more abstract features.
	In Fig.2 the filters for the toy data represent parts of corners, as that is the only thing that differentiates a circle from a square. This is further corroborated by the activations shown in Fig.3, where the top left and bottom right corners have highly differing activations on different sides.

\begin{figure}[]
\centering
\includegraphics[width=\columnwidth/2,angle=90]{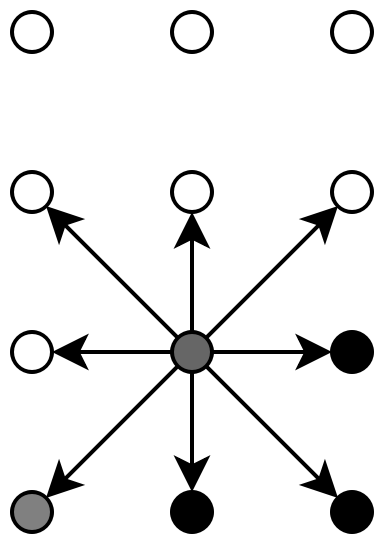}
\caption{A section of an image as a point cloud.}
\end{figure}

After several layers of these generalized convolutions have been applied, we finally apply one more, where the "query point" is (0,0,0) and the nearest neighbors are all of the points in the penultimate layer. The conventional convolutional analog would be a fully connected layer, which can be thought of as a convolution taken at a single pixel, with the filter being the same size as the input image. Alternatively, if we are looking to segment the image semantically and label each point, we can apply this final layer to every point, in the exact same way that fully convolutional neural networks\citep{DBLP:journals/corr/ShelhamerLD16} do. Fig.5 shows the full algorithm in the more intuitive computation graph form that it takes when built for computation. It includes features that are discussed in detail in the next section, notably a form of striding.

\begin{figure*}[ht!]
    \centering
    \begin{subfigure}[t]{0.32\textwidth}
        \centering
        \includegraphics[width=\linewidth]{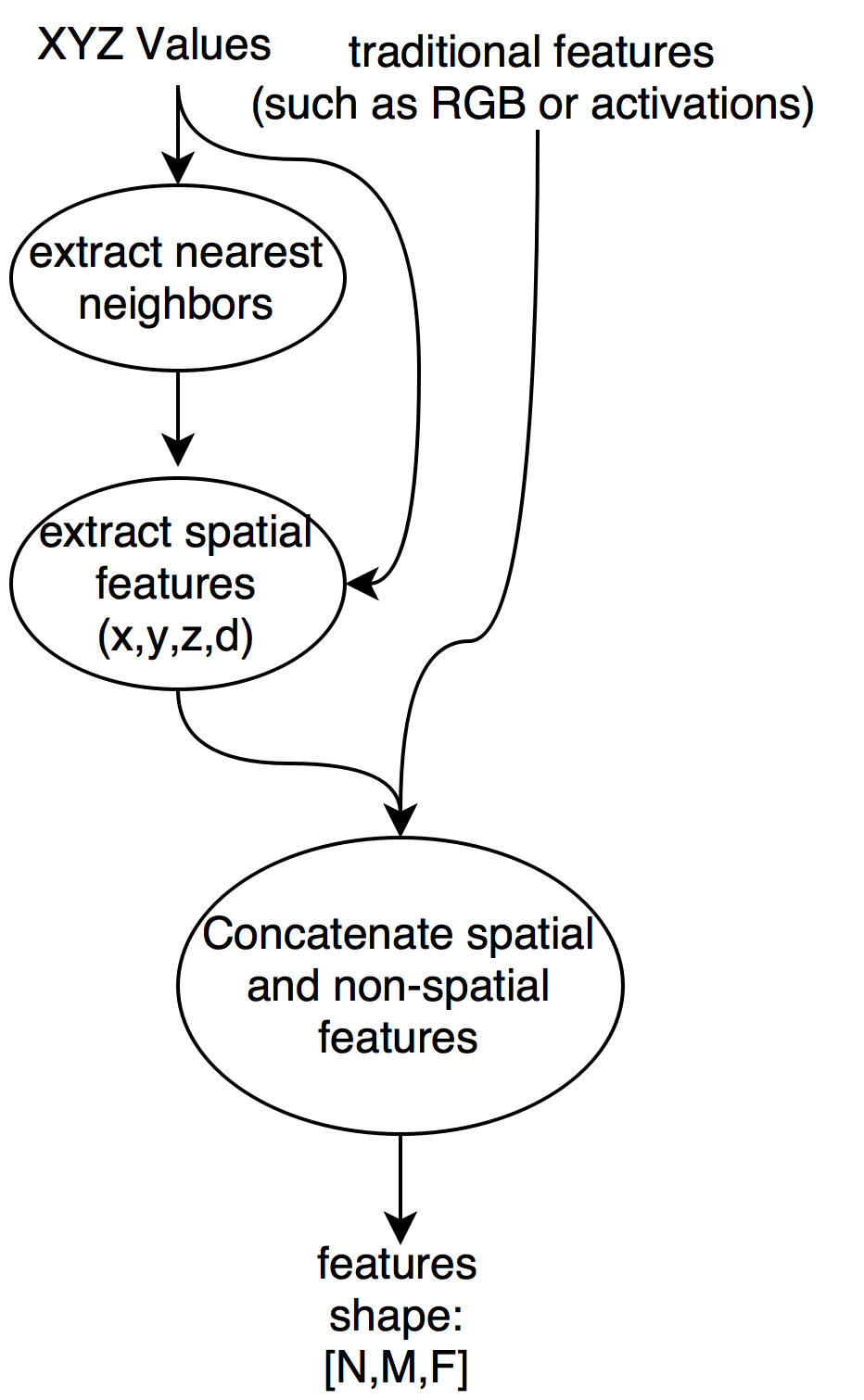}
        \caption{Spatial Feature Extractor}
    \end{subfigure}
    \begin{subfigure}[t]{0.32\textwidth}
        \centering
        \includegraphics[width=\textwidth]{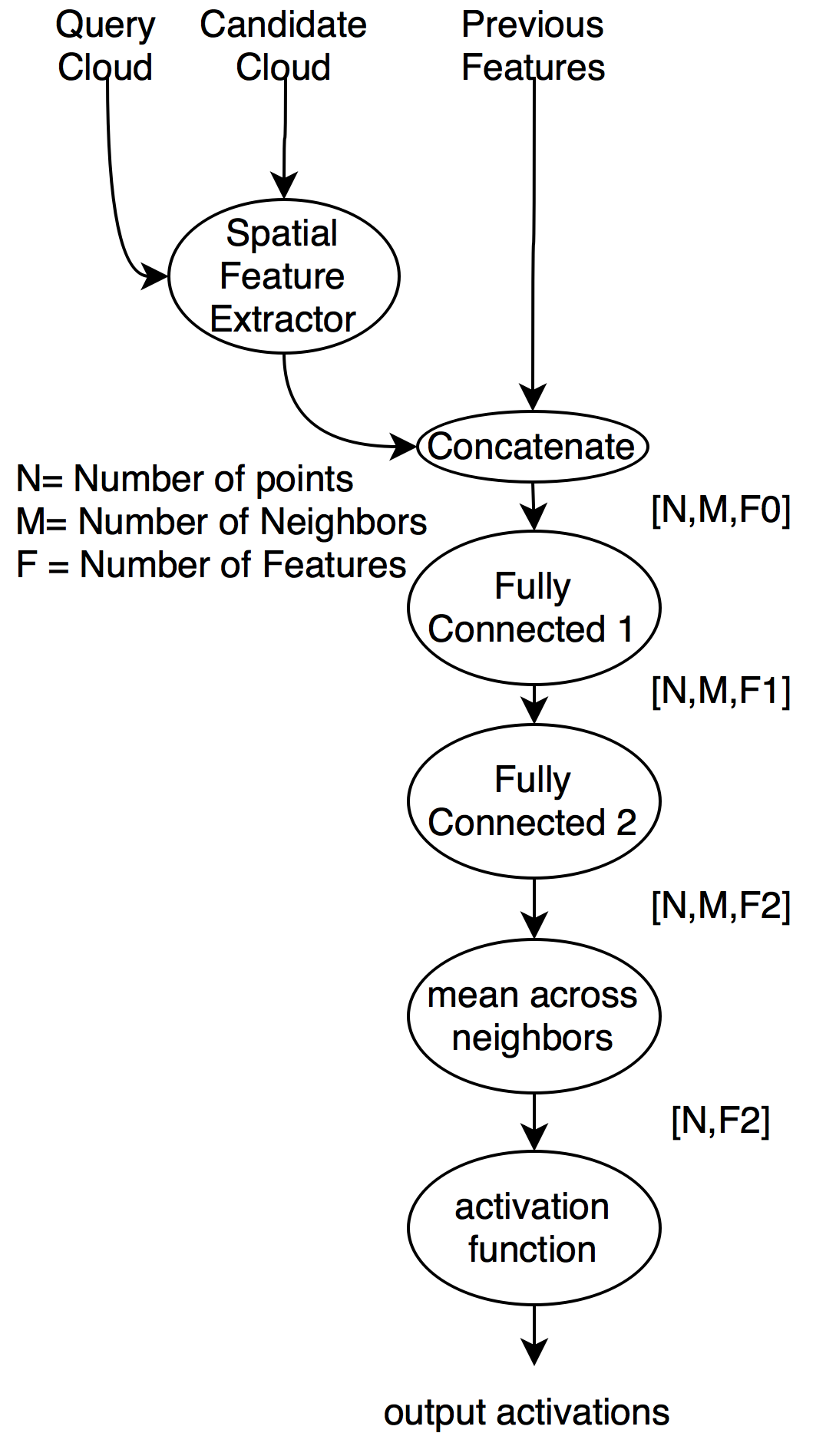}
        \caption{GenConv Module}
    \end{subfigure}
    \begin{subfigure}[t]{0.32\textwidth}
        \centering
        \includegraphics[width=\textwidth]{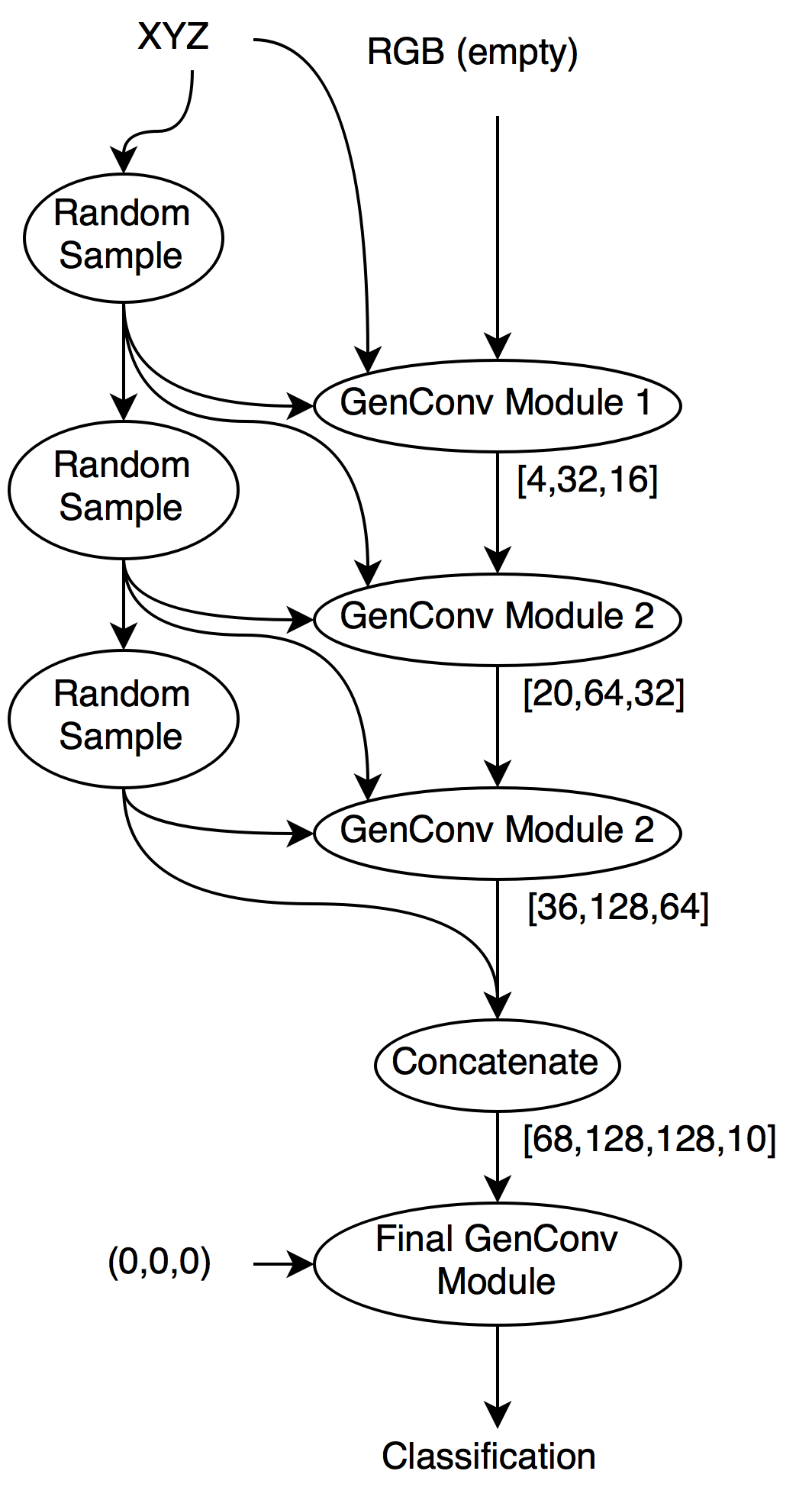}
        \caption{Final Model used for the ModelNet10 Dataset.}
    \end{subfigure}
    \caption{Core components of generalized convolution. Bracketed numbers in in subfigure c represent the sizes of internal neural network layers for their modules.}
\end{figure*}

\section{Analogies to Already Existing Systems}
We claim this technique is a "generalized" convolution, thus it must be able to represent a conventional convolution. If we take a 2D image and split it into points on an integer grid as in Fig. 4, we can apply both types of convolutions to the same data. A conventional convolution outputs the sum of a function applied once to each point within a certain receptive field. This function $f(\Delta x,\Delta y,a_{\Delta x, \Delta y})$ is defined piecewise for each possible spatial relationship a point in the receptive field might have to the point at which the convolution is evaluated. That is $f(\Delta x,\Delta y,a_{\Delta x, \Delta y}) = a_{\Delta x,\Delta y}*\theta_{\Delta x,\Delta y}$. This is reasonable, and even efficient, because in the case of a normal image, each point is guaranteed to be an integer $\Delta x, \Delta y$ away from the point at which the convolution is evaluated. However, if such an assumption cannot be made, the learnable piecewise function must be replaced by a continuous function. We define a new $f(\Delta x,\Delta y,a_{\Delta x, \Delta y})$ to be a small neural network. While evaluating a function is slower than simply fetching the weight values, this allows us to apply the model to point clouds directly.

The visualization process for convolutional neural networks also needs to be generalized. While we currently simply view an image of the filters, this is actually a proxy for information with more meaning. Because the output of a convolution is the sum of each $\omega*x$, we can think of each pixel as "contributing" to the output of the convolution as a function of the weight. Thus, our current visualizations of convolutional filters show "how much would a pixel contribute to the image if it was constrained to a value of 1. We, of course, skip this step as it is a multiplication by 1, and simply show the weights themselves.

In the case of generalized convolutions, we need to apply the concept of a points "contribution" fully. We create a dummy point $\Delta x, \Delta y$ that we are curious about, and attach conventional features of 1 to it. We then pass it through the generalized convolution. The output is the "contribution" of a point in that location relative to the center of the filter. Doing this iteratively for a grid of points yields a visualization that is similar to those of conventional convolutions. The granularity of this grid however, is flexible due to the continuous nature of the filter. Thus, very high resolution visualizations are possible.Fig. 2 illustrates the filters for our toy network sampled at a resolution of 51px by 51px, and Fig. 6 visualizes the 3D filters for the first convolution on the modelNet10 set.

\begin{figure*}[ht!]
\centering
\includegraphics[width=\textwidth]{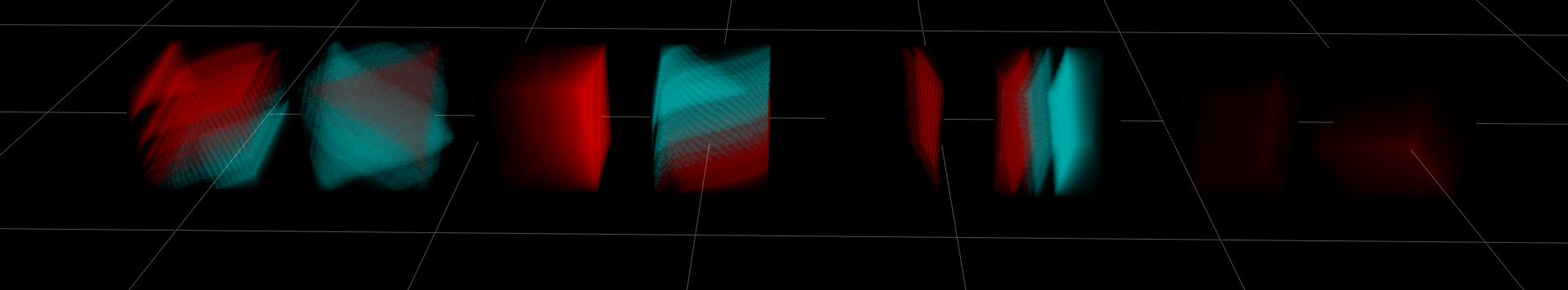}
\caption{Filters learned on the ModelNet10 dataset}
\end{figure*}

This similarity shows itself in the properties of generalized convolutions,  notably that deeper networks with smaller filter sizes seem to be most effective per parameter.

These convolutions also closely mirror graph convolutions\citep{DBLP:journals/corr/KipfW16}, mostly because they can be configured to emulate the concept. If we were to use the same set of points as both the query set and the candidate set for neighbor extractions (As we do in the simple variant of the algorithm above), we would effectively create a directed graph of nearest neighbors. Applying a generalized convolution to this graph would cause each "node" (point) to store a vector representing both its features and its relationship to those nodes it is connected to. In order to apply generalized convolutional layers to actual graphs, all one needs to do is omit the nearest neighbor step and use the locality data already present in the graph.

As things stand we do not limit the model to a graph representation, however, as we can omit arbitrary points and introduce new points at every layer. This is an integral part of the process in fact, and can be viewed as a form of striding. In the final configuration of the model, we sample half of the candidate points for use as query points at every layer, resulting in input sizes of 1000, 500, and 250 points for each layer.

\section{Complexity Analysis}

The primary motivation behind the creation of this new layer was inefficiency in applying conventional architectures to the point cloud problem. A conventional convolution has the complexity O($N_x*N_y*N_z*n_x*n_y*n_z$) where $N$ represents input dimensions, and $n$ represents the filter dimensions. This makes conventional convolutions seem to be $O(N^3)$ with respect to the size of the input, however the situation is much more complex when point clouds are involved because the size of individual dimensions in 3D space is not rigidly correlated to the size of the input point cloud. The issue at hand is that of sparsity/density. Because the conventional matrix oriented definition of sparsity does not apply to continuous spaces, we will define sparsity as the volume per point. If we assume that sparsity is constant for all sizes of inputs, the complexity class of a conventional convolution is actually only $O(N)$ where N is the number of points. That is to say, if the number of points were to octuple, but the input were to maintain its sparsity, the product $N_x*N_y*N_z$ would also only octuple. The assumption that sparsity remains the same is not one that we can make however, as the "sparsity" depends on the level of detail that one wishes filters to learn. This means that the complexity class is actually $O(N*S(N))$ where $S$ represents the function that best approximates sparsity. We don't define this function because it is situational, and likely does not approximate well to any common functions in many cases. However, if the level of detail remains the same, and the scale of the problem grows (classifying a cloud as a sedan vs finding a sedan in a scene), sparsity increases. There is also the issue that the level of detail in real world scans varies across the image due to varying distances from the sensor. This requires that conventional convolutions work with smaller blocks, increasing sparsity further. Applying a conventional convolution sparsely to combat this fact is challenging because convolution has a dilatory effect on tensors, where sparse tensors quickly become dense after very few convolutions. This means that when converting point clouds into occupation grids, one needs to keep sparsity, and thus resolution, low to maintain an input of reasonable size, losing information in the process.

In contrast, the computational complexity of our layer is dependent only on the size of the input, and is not affected by sparsity. This results in a complexity class of $O(Nlog(N) + O(N*n)) -> O(Nlog(N))$, where N is the number of points and n is the number of parameters in the layer. The $O(Nlog(N))$ component arises from the need to search for nearest neighbors through the use of a KD-tree and should in practice not be noticeable due to the large size of $n$ in most models. This low complexity comes at the cost of requiring that the filter in regard to spatial features be calculated for each neighbor separately, whereas the conventional model essentially has a "lookup table" with weights for each spatial relationship that is possible. This all means that our method is more efficient than conventional convolution when sparsity is above a certain threshold, i.e for $N$ such that  $S(N)>Log(N)$. We show empirically that this is the case for the ModelNet10 dataset.

\begin{table*}
	\centering
	\resizebox{\textwidth}{!}{
		\begin{tabular}{ |c|c|c|c|c|c| }
\hline
Type & Method & Augmentation &  Size & ModelNet40 & ModelNet10\\
\hline
Ensemble Volumetric & VRN Ensemble\citep{DBLP:journals/corr/BrockLRW16} & 24x rotations & \textbf{90M} & \textbf{95.54}\% & \textbf{97.14}\%\\
Ensemble Vol/Multi & FusionNet\citep{DBLP:journals/corr/HegdeZ16} & 60x rotations & 118M & 90.80\% & 93.11\%\\
Ensemble Multiview & MVCNN\citep{su15mvcnn} & 80 views & $\sim$138M** & 90.10\% & \\
Ensemble Multiview & Pairwise\citep{DBLP:journals/corr/JohnsLD16} & 12 views &  & 90.79\% & 92.80\%\\
\\ \hline
Single Volumetric & 3DShapeNets\citep{DBLP:journals/corr/WuSKTX14} & 12x rotations & 38M & 77.32\% & 83.54\%\\
Single Volumetric & VRN\citep{DBLP:journals/corr/BrockLRW16} & 24x rotations & 18M & 91.33\% & 93.61\%\\
Single Volumetric & VoxNet\citep{VoxNet} & 12x rotations & 0.92M & 83.00\% & 92.00\%\\
Single Volumetric & ORION\citep{DBLP:journals/corr/AlvarZB16} & 12x rotations & $\sim$0.92M* &  & 93.80\%\\
Single Volumetric & PointNet\citep{PointNets} &  & 90M &  & 77.60\%\\
Single Volumetric & Subvolume Sup.\citep{DBLP:journals/corr/QiSNDYG16} & $\sim$20 rot+elev & 16M & 87.20\% & \\
Single Volumetric & LightNet\citep{3dor.20171046} & 12x rotations & 0.3M & 86.90\% & 93.39\%\\
Single Multiview & DeepPano\citep{DeepPano} &  &  & 82.54\% & 86.66\%\\
Single PointCloud & PointNet\citep{PointNets} &  & 8M & 89.20\% & \\
Single PointCloud & Kd-Networks\citep{DBLP:journals/corr/KlokovL17} & translate+scale & 2M & \textbf{91.80}\% & \textbf{94.00}\% \\
Single PointCloud & Generalized Conv & None & \textbf{0.04M} &  & 92.40 $\pm$ 0.4\% \\
\hline
		\end{tabular}}
	\caption{A comparison of relevant performance and parameter counts of various architectures on the ModelNet datasets. Blank spaces signify data could not be found. "*" indicates based on VoxNet structure. "**" indicates based on VGG-M structure.}
\end{table*}

The space complexity of Generalized Convolutions is also favorable, as conventional 3D convolutions grow by $O(N^3)$ where N is the resolution in X,Y, and Z. Generalized Convolutions are bound only by the complexity of the feature they need to represent and the number of points one wishes to include, thus ($O_{parameters}$ * $O_{neighbors}$), and the "size" of the convolution can be finely tuned independently of the "resolution". In practice, this yields noticeable gains as seen in Table 1. 

\section{Implementation and Real Performance}
Using this new type of layer we were able to achieve a 92.4 $\pm$ 0.4\% classification accuracy on ModelNet10\citep{DBLP:journals/corr/WuSKTX14}, competitive with other single model implementations, with only 41,888 parameters after training for 20 epochs (approximately 40 minutes on an AWS g2.2 instance). We intentionally compare our results only to those of other single model publications, as \citep{DBLP:journals/corr/BrockLRW16} has already shown that ensembling can increase performance on this type of data. 

The configuration of the general convolution layers is shown in Fig.2c. Leaky relu activations were used everywhere except for the final softmax activation. The model was trained with a batch size of 1, a learning rate of 0.005 using Adam\citep{DBLP:journals/corr/KingmaB14}, no l2 loss, and with a learning rate decay of 0.96 every 500 samples. Our implementation uses Tensorflow and is available at https://github.com/ThaHypnotoad/GeneralConvolution.

Note in this diagram that there are currently no fully connected layers involved. Instead the final layer is simply a wide and deep generalized convolution. 

\section{Future Work}
Notably absent from this architecture is any form of pooling. While there are plenty of algorithms for downsampling point clouds, it may not be immediately necessary to implement pooling explicitly, as it has been shown that in conventional convolutional neural networks, pooling can be replaced by convolutions with greater stride \citep{DBLP:journals/corr/SpringenbergDBR14} and applied our resources elsewhere.

Currently, the architecture utilizes a fixed-sized nearest neighbor search instead of a variable sized neighborhood search. This design decision is to avoid non-uniformly sized arrays, which are difficult to express in matrix-matrix operations. In addition, a nearest neighbor search allows the model to apply appropriate radii to areas of varying sparsity. We have yet to look into the implications of changing the behavior of feature extraction in this way. The current implementation of this technique also omits many standard machine learning techniques such as data augmentation, ensembling, batching (due to again, an issue with jagged arrays) and thus batch normalization. We suspect that properly re-implementing these techniques for this type of model will yield similar benefits to those conventional deep CNN models experienced.

As previously mentioned, because we are working in a continuous space, it is possible to not use any of the original point cloud for the query and select entirely new points not in the original input. We do not however, instead constructing the query cloud from a sample of the input to the layer. Were we to develop a computationally cheap algorithm for selecting or generating salient points, we could significantly increase the extent to which we apply striding and thus speed. We plan to look into such algorithms in the future.

Finally, this type of model is applicable to classes of machine learning problems where relationships between data points are relevant to the individual data points themselves. Graphs, for example, are such a class and we intend to test Generalized Convolutions on them in the future. "Spatial features" can be any feature that indicates a relationship between features instead of an absolute feature. It just happens to be that the most obvious such features in a point cloud are spatial relationships.

\section{Conclusion}
Generalized Convolutional Neural Networks achieve competitive performance among other single model architectures using significantly fewer parameters, and scale well in the case of sparse point clouds. They also have the advantage over voxel based methods in that they natively work with point clouds, requiring no voxellization in real time applications.

\section{Acknowledgements}
This work is supported by Cylance Inc.

\bibliographystyle{IEEEtranN}

\bibliography{references}
\end{document}